# Syntax-aware Hybrid prompt model for Few-shot multi-modal sentiment analysis

Zikai Zhou[a,#], Haisong Feng[b,#], Baiyou Qiao[a,b,*] , Gang Wu[a], Donghong Han[a]

[a]*School of Computer Science and Engineering, Northeastern University, Shenyang, 110819, China*

[b] *School of Informatics, Xiamen University, Xiamen,361105, China*

# *These authors contributed equally to this work*

\* *corresponding author: qiaobaiyou@mail.neu.edu.cn*

**Abstract**

Multimodal Sentiment Analysis (MSA) has been a popular topic in natural language processing nowadays, at both sentence and aspect level. However, the existing approaches almost require large-size labeled datasets, which bring about large consumption of time and resources. Therefore, it is practical to explore the method for few-shot sentiment analysis in cross-modalities. Previous works generally execute on textual modality, using the prompt-based methods, mainly two types: hand-crafted prompts and learnable prompts. The existing approach in few-shot multi-modality sentiment analysis task has utilized both methods, separately. We further design a hybrid pattern that can combine one or more fixed hand-crafted prompts and learnable prompts and utilize the attention mechanisms to optimize the prompt encoder. The experiments on both sentence-level and aspect-level datasets prove that we get a significant outperformance. Then we conduct
 an ablation study and a series of further analysis

## 1. Introduction

Multi-modal sentiment analysis (MSA) has gained popularity among researchers in recent years because of its wide practicality in e-commerce [1], social media [2], and human-computer interaction [3]. Previous studies commonly cast MSA into two kinds of sub-tasks: sentence-level sentiment detection with given representations of modalities and aspect-based sentiment analysis (MABSA) with given aspect terms and accompanying multi-modal pairs. In this paper, we mainly adopt textual and visual modalities to execute the tasks above mentioned and evaluate our work.

However, most existing approaches ignore the fact that labeled multi-modal datasets are large and expensive. Hence, the most practical solution should be to investigate dew-shot learning strategies that have the ability to perform relatively well with a low-resource training set, such as LM_BFF [4] in text-based tasks, etc. Prompt methods have been popular for NLP tasks since PET [5]'s proposal, for they make pre-trained models execute new tasks with low resources or even no data for training. Besides, the recent study PVLM [6] in MSA has demonstrated that prompt tuning fits not only textual but multimodal scenarios. PVLM separately tests two main types of prompt methods: fixed and unlearnable prompts and learnable prompts for few-shot multimodal sentiment analysis (FMSA) and few-shot multimodal aspect-based sentiment analysis (FMABSA). However, for the fixed prompt, the best choice of PVLM in most cases, it's well known that the fixed prompt has one disadvantage: it's hard to craft the optimal prompt by humans. As shown in the result of PVLM, each hard prompt gained the best performance in the partial dataset. Moreover, we find that the learnable prompt also achieves and even outperforms the hard prompt on a special dataset.

So we have reason to believe that an effective fusion of two prompt-based methods mainly used to enable prLM (pre-trained language model) to perform better for FMSA and its design is exchanging simultaneously.

To solve the above challenge, we propose the hybrid syntax-enhanced prompt model for sentiment reorganization of text and vision double-modalities, relying on a few trainings. Our method incorporates fixed and learnable prompts to improve the multimodal few-shot effect and transform FMSA and FMABSA tasks into a language modeling problem with masks that return the probability of being designated sentiment words (verbalizers) based on given hybrid prompts. Specifically, we first conduct a hybrid prompt pattern to merge one or more hard prompts with learnable prompt tokens appropriately, which do not simply merge two single hard prompts or hard and soft prompts but contain hard prompts and learnable (soft) tokens together. We find that the prompt encoder of PVLM performs unsatisfactorily. So, with PLVM's as the backbone, we further optimize it into an interactable and syntax-aware reference where biaffine attention is introduced to referencing the syntax knowledge and SDPA attention to enriching the semantic information through interactively transferring representation among learnable tokens. Our main contributions can be summarized as follows:

- We propose a novel multimodal hybrid syntax-aware prompt model for both FMSA and FMABSA tasks.
- We designed the hybrid pattern to merge fixed prompts with learnable prompts to make it more robust.
- We propose a syntax-aware inference unit wrapped with the biaffine and scaled dot product attention (SDPA) mechanisms to encode a learnable prompt for FMSA.
- We execute experiments on four multimodal sentiment datasets and outperform the existing method significantly.

## 2. Related Work

### 2.1. Sentiment Analysis and Multi-modal Sentiment Analysis

The core task of SA is to predict the sentiment polarity of given sentences. It had been a popular direction for research and business in NLP (neutral language processing) as early as 2010. There are a number of review books and papers covering a wide range of early methods and applications. [7,8,9]. Aspect-based Sentiment Analysis (ABSA) as a subtask of SA, proposed to predict the sentiment polarity for aspects, has received high attention. The research relatively includes aspect-opinion co-extraction [10–14], aspect-opinion pair extraction [15–16], and aspect-opinion-sentiment triple extraction [17–20]. Recently, the approaches for the ABSA task mainly utilized tree and graph structures based on language syntax knowledge [20–24].

Similarly, in a multi-modal scenario, at the sentence level, multi-modal sentiment analysis (MSA), as another subtask of SA, on text-image pairs aims to recognize the sentiment according to the given text and corresponding image. [25] publish a cross-modal emotion detection dataset on textual and visual modality and design a multi-view network with attention to the features of visual objects, visual scenes from images, and textual representation. [26] proposes a Coupled-Translation Fusion Network, achieving multi-modal information interaction through coupled learning to ensure robustness when modalities miss. [27] find that automatic speech recognition may cause errors and hurt the prediction task, based on which they propose to refine the erroneous sentiment words by predicting them from other modalities. At the aspect level, different from text-based tasks, it's acknowledged as challenging to fuse textual and image information. As the pioneer in MSA, [28] collected a benchmark dataset in Chinese from a digital product feedback online platform and designed a memory network gathering multi-interaction information to fuse representations from both textual and visual modalities. More recently, [29] contributed two datasets with open-ended aspects from Twitter for MABSA and effectively combined both modalities by leveraging BERT as the backbone. In the same period, Yu proposed a fusion network with target-aware attention, solving target-based sentiment reorganization in both textual and multi-modal modalities. Then [30] first utilizes cross-modal triplet extraction to assist in the multi-modal aspect-level sentiment classification task.

However, all the above approaches depend on large, multi-modally labeled data, which has been difficult to collect and annotate. Different from them, our proposed method, which solves the few-shot multi-modal sentiment analysis and aspect-based sentiment analysis tasks, avoids the necessity of a large training set by simply relying on a few training sets.

### 2.2. Few-shot learning and prompt-based method

For the reason mentioned above, so that the recent research starts to pay attention to the few-shot learning for SA, we list the alternative works in relevant sentiment analysis tasks we find: [6] conducts the Trans-Modal Information Fusion Model PVLM based on a few samples in multimodal sentiment analysis; [31] transforms sentiment classification into a text generation task and tests on a few shots and the full shot, achieving significant improvement. In AP [32], based on the additive attention mechanism, an adaptive encoder is constructed, which is trained by large-scale data sets, and transform learning is applied to a few-shot task. Within the best scope of our finding, only PVLM can be referred to and followed for the multi-modal sentiment task, which utilizes the prompt-based method for FMSA for the first time and proves that prompt tuning replacing fine-tuning is significantly beneficial.

Prompt tuning, originating from prompt learning (prompt-based learning), Since the early period, prompt learning was just integrating the constructed template into the original text, but different manual discrete templates have proven to have a great impact on the results. Later, the continuous prompt method was proposed to avoid manual construction, which achieved similar performance or even better than the manual template. Essentially, prompt learning is an efficient way to make processing closer to natural language, converting downstream tasks into corresponding PrLM self-supervised learning tasks. Recently, Chen et al. (2021) [33] solved the relationship extraction task with the Mask Language Model (MLM)—BERT [39]. Sun et al. (2021) [34] applied prompts to the downstream task, the next sentence prediction (NSP). Moreover, as the pre-training model fits successfully on NLP tasks, the prompt-based pre-training model can perform well on few-shot tasks, too. On many text-based tasks, [35] utilizes auto-constructed prompts for text categorization. [36] introduce prompts, significantly improve the result of text regression on few-shot datasets, and construct prompts to assist text generation tasks. [37] use prompts to improve the generalization performance of the pre-training language model on different one-shot tasks.

Based on these approaches, we propose a hybrid prompt that combines both a hand-crafted template (the hard prompt) and a learnable template (the soft prompt) to sufficiently use prompt-based methods. Besides, we employ the attention mechanisms for gathering syntax information and semantic representation to optimize the prompt encoder of PLVM and further explore the prompt tuning's potential for FMSA.

## 3. Proposed Methods

Figure 1. The $P$ denotes the pseudo-tokens, with the subscript meaning the length, the Hybrid prompts pattern combines

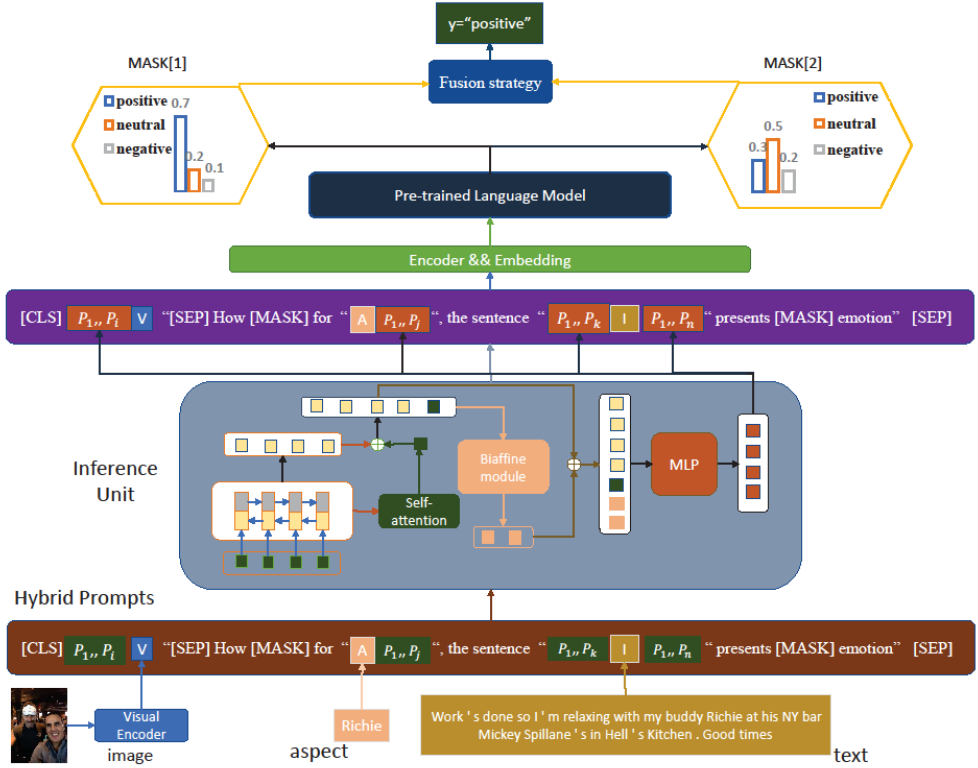

the input from original modalities and prompts, then the Inference unit encodes and inference for pseudo-tokens(colored in brown) and utilizes the output to replace the original tokens(colored in green), the encoder and embedding depend on the pre-trained model, the [MASK]'s are shown as MASK[1] and [MASK]2, the prediction scores of them are different. In this case, right prediction(MASK[1]) and wrong (MASK[2]) are utilized by the Fusion strategy, merging directly or choosing. Here the result after merging is "positive" with the highest prediction scores, too.

We first introduce how our proposed framework catches and processes the representation from two modalities with the designed hybrid patterns in cross-modality tasks.

For the vision modality, following the PLVM, we leverage the NF-RestNet to extract and eject the neural vision representation into the textual embeddings and format the process as follows:

$$V = WiPool(ResNet(X)) + bi$$

$$V^{\sim} = reshape(V) = [v^1, \dots, v^j, \dots, v^{N_i}], \qquad v^j \in R^{d_t}$$

where $V \in R^{d_{nt}}$ , $W_i \in R^{d_t \times d_{nt}}$, $b_i \in R^{d_{nt}}$, nt $= d_t \times N_i$, $N_i$ that is defined to limiting the number of tokens representing original vision features. $X$ denotes the input of image modality and $d_t$ refers to the dimension of textual embedding for the PrLM.

For textual modality, we optimize the prompt encoder of p-tuning by attention mechanisms. In detail, we utilize the SDPA to enrich the semantic information and biaffine attention to understand the syntax information helpful to reference (We explore the combination strategy of two attention mechanisms in further analysis and employ the best framework) and not simply gather the representation from both modalities alternatively, but also supplement the hand-crafted hard prompts.

The figure "?" shows the framework of our proposed method. In this figure, we take an instance with two [MASK] tokens in hybrid pattern for aspect-level tasks.

### Task definition

In this paper, we assume that it is possible to obtain a pre-trained language model M, such as BERT, and that we aim to tune it on the multi-modal sentiment analysis (FMSA) task and the multi-modal sentiment analysis (FMABSA) task with just a few training sets.

For sentence-level MSA (FMSA), given a multi-modal sample $S$ and emotion labels L, it contains an image and a corresponding text $I = i_1, \dots, i_n$, in which n denotes the length of the tokenized sentence. Our purpose is to recognize that the classification of the sentiment expressed by both the text and image corresponds to "positive", "neutral" or "negative" as the space of L for 3-classes or "positive", "negative," for 2-classes.

For aspect-level MSA, given a text-image sample $S$, $I$, image, $L$, similar to MSA, with a phrase or word as aspect $A = a_1,...,a_m$ especially, in which m denotes the number Our goal is to predict which space it will most probably fall into among the spaces of the $L$ with respect to $A$.

**Multimodal-aware prompt-based tuning**

Multimodal models proposed by related approaches are prone to being infectious because of overfitting. An effective strategy for solving this problem is prompt tuning instead of fine-tuning. We conduct the prompt-based method on text modality and embed a visual model, which represents the original input from the image feature space. Given a pre-trained language model (PrLM) L, prompt tuning directly tasks the L with the natural language template or pseudo-tokens for adaptation learning.

**Hybrid pattern for prompt-based tuning**

For instance, on a text sample, we can design the pattern, transforming the classification task of three polarities (i.e., positive, neutral, and negative) into a cloze and mask-filled task. It can be formatted as follows: For $I$ (e.g., "a delicious food"), $I_{hard} = T (I)$ :

$$I_{hard} = [CLS]I \ how \ [MASK].[SEP]$$

$$I_{soft} = [CLS]I \ [unsed][unsed][MASK][unsed][unsed][SEP]$$

Let $L$ conjecture whether it is more possible to fill in "delicious" (positive) or "bad" (negative) for "mask." In our hybrid pattern, they are combined as follows:

$$I_{hybrid} = [CLS]how \ [MASK]. \ I[unsed][MASK][unsed][SEP]$$

when hard templates merge, as follows:

$$I_{hybrid} = [CLS]how \ [MASK]. It \ is \ [MASK].[unsed][MASK][unsed]$$

For aspect-level (single text modality), it is similar to the previous patterns except for the separate joining of aspects. $I_{hybrid} = T (I, A)$, Formally,

$$I_{hybrid} = [CLS]how \ [MASK]. For \ A, \ it \ is \ [unsed][MASK][unsed][SEP]$$

In a multi-modal task, we merge the pseudo-visual tokens $V = v_1,,, v_j$, which work by generating the feature space from the visual modality into the above patterns. By doing this, we combine the text and vision. So that the prompt method working for PrLM is successfully applied to both modes in the FMSA task. For sentence-level MSA:

$$I_{hybrid} = [CLS]V[SEP]how \ [MASK]. for \ A \ [unsed]I[unsed] \ [MASK] \ [[SEP]$$

where $j$ denotes the length of $V$. Similarly, for aspect-level, aspect terms $A$, original text $I$, and vision tokens $V$ compose into $I_{hybrid} = T (I, V, A)$.

$$I_{hybrid} = [CLS][unsed][unsed]Vj \ [SEP]how \ [MASK]. for \ A, [unsed]I[unsed] \ [MASK] \ [SEP]$$

It is worth noting that there are some considerations due to the specific design. We give the form for reference, in which $i$, $j$, $k$, $m$, $n$, $y$, and $z$ all represent the length of the given prompts $p$ .$H_{prompt} = $ [for "" the sentence is [MASK]] , where the characters $I$ and $V$ mean the input of text and image modality.At the sentence level (double fixed templates)

$$I_{hybrid} = [CLS][unsed][unsed]Vj \ [SEP]how \ [MASK]. for \ A, [unsed]I[unsed] \ [MASK] \ [SEP]$$

Ensure that $z$ is not 0 to ensure the functionality of the visual modality.
At the aspect level (single fixed template, for instance; double fixed templates similarly)

$$I_{hybrid} = [CLS]p_i \ how \ p_j[MASK] \ p_k \ the \ sentence \ p_m \ I \ p_n \ is \ [MASK]p_y \ [SEP] \ p_z \ V \ [SEP]$$

Ensure that $z$ is not 0 and $j$ is 0, avoiding destroying the aspect terms.

We show examples in table $$ for a detailed description with red color for Inappropriate setting.

| | | |
|---|---|---|
| {[unsed]} for A the sentence I is [MASK] {[unsed]} [SEP] {[unsed]} V [SEP] | Right instance | ✓ |
| {[unsed]} for ▮▮▮▮▮ A the sentence I it is [MASK] {[unsed]} [SEP] {[unsed]} V [SEP] | Destroy aspect terms | ✗ |
| {[unsed]} for A the sentence I is [MASK] {[unsed]} [SEP] ▮▮▮▮▮ V [SEP] | Ignore the vision modality | ✗ |

Table 1. In the table,{unsed} means a simple prompt with length 1 for example, *A* represents the aspect terms, *I* , *V* denote original input from textual and image modality.

**Syntax-aware Inference module based on multi-attention**

**Syntactic depende**ncy

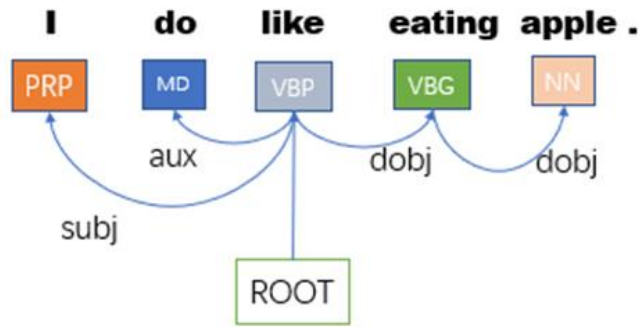

Figure 2: A dependency tree parse for "I do like eating apple", including the vertexes representing word, root, and relationship edges from child to parent, which are labeled by relationship types.

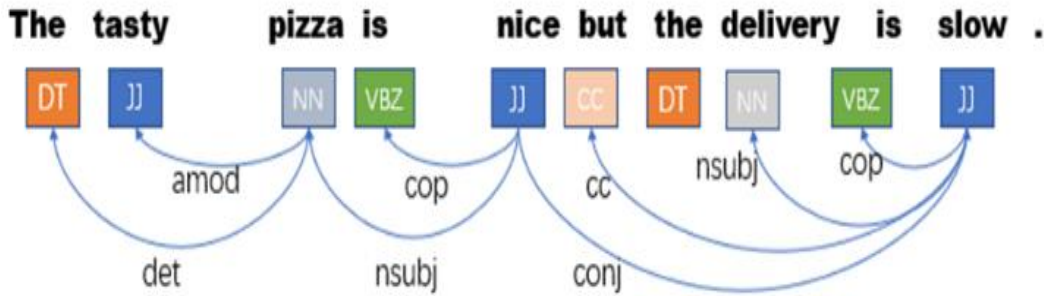

Figure 3. A sentence "The tasty pizza is nice but the delivery is slow "with its dependency tree for aspect-based sentiment analysis task, where the aspect terms are highlighted in yellow and words expressed different emotion polarities in red and green.

We utilize examples to display the dependency in textual mode and how it inflects the SA and ABSA tasks.

In sentential affective analysis, e.g., Figure 2, the sentence is parsed as a tree by dependency. As shown, the word "like" repressing the emotion is parsed as the "root" of a tree and has types of relationships (detailly referring to http://www.universaldependencies.org) with "i","do", and "apple" acting as the child of "eating" in edge "eating->apple" because "apple" depends on "eating."

For aspect-level, take Figure 3 as an instance: for edges that contain words ("tasty" "pizza"), "tasty" and "pizza" are part of the aspect term "tasty pizza". Similarly, for the edge "nice->pizza," "pizza" acts as an opinion target of "nice" and is endowed with positive emotion. Based on this, it is understandable that the finding of dependency between "tasty" and "pizza" is helpful to enhance the aspect term and recognize the sentiment for "pizza". Likewise, the discovery of the edge between nice and pizza rather than understanding "pizza" with "slow" is necessary and effective. In short, the discovery of dependency and the learning of task-dependent word representations for SA and ABSA both have positive effects on assistant performance.

**Biaffine Attention Mechanism**

The biaffine attention mechanism, proposed for neural dependency parsing, has been proven effective in syntactic dependency parsing. Besides, it is widely used in text-based explicit dependency trees and GCN-based models. It is understandable that learning the syntax information makes the model's understanding of the representation of natural language more similar to that us human.

In previous approaches, they labeled each word of a sentence with tags and computed the cross-entropy loss. Unfortunately, the pseudo-tags embedded in the learnable sequential prompt method mean that they cannot carry explicit tags. However, it is well known that cue-based learning makes input contain more natural representations or learn more natural representations adaptively. Based on this, they have improved on few-shot and even zero-shot, so we think that the introduction of the biaffine attention mechanism to learn grammar representation is helpful in the few-shot task.

Before being combined with the original text, Biaffine learns the syntax dependencies that pseudo-tag locations should learn adaptively. Specially, we feed the representation information integrated by BI-LSTM layers and self-attention as double-channel inputs into the Biaffine Attention Module, which extracts the syntax representation and enriches the representation of textual and image modalities. In the biaffine attention module, a multi-layer perceptron (MLP) is utilized for the input of channels to reduce the dimensions of inputs, followed by two deep bilinear attentions. The process formats as follows:

$$h_i^f = MLP_f(h_i)$$

$$h_j^s = MLP_s(h_j)$$

$$g_{i,j} = h_i^{f\ T} U_1 h_j^s + U_2 \left(h_i^f \oplus h_j^s\right) + b$$

$$r_{i,j,k} = \frac{\exp(g_{i,j,k})}{\sum_{l=1}^m \exp(g_{i,j,l})}$$

$$R = Biaffine(MLP_f(H), MLP_s(H))$$

The score vector $r_{i,j} \in R^{1 \times m}$ formats the relations between $w_i$ and $w_j$; $m$ means the types of relations, and $r_{i,j,k}$ denotes the score of the $k$-th relation type for pair $(w_i, w_j)$. The adjacency vector $R \in R^{n \times n \times m}$ (because the input of channels has the same dim $n$ shown sentence length) models the relation between words. $U_1$, $U_2$, and $b$ are learnable weights and biases. $\oplus$ denotes concatenation. Eq. (16) collects process of Eqs. (14) to (16)

**Interaction-based self-attention (SDPA)**

To integrate with fixed manual templates and fit the few-shot task of multi-modal sentiment analysis (FMSA), we introduce the SDPA mechanism from Transformer. With the assistance of scaled dot-product attention, the inference unit can better capture the semantic information of an input sequence by learning the interaction information in the input sequence based on the similarity among them.

Moreover, when integrating SDPA into the inference unit, the trainable prompt template encoder remembers and extracts more semantic information about the hard template. It is suitable and helpful for a mixed prompt strategy.

**Loss computation**

In our approach, we define mappings $\varphi: Y \rightarrow V$, where $Y$ represents a label space of labels in a data set, and $V$ represents a collection of separate vocabularies in the MLM model. For each triplet of visual input $V$, text $O$, and aspect $A$, $I$ denote the original input for each sample in the pattern, which contains pseudo-tags, human-built prompt prompts, and one or more [MASK] tokens. We can use the sentiment classification task to mask cluster problems in MLM and compute the probability of predicting class $y \in Y$ as:

$$p(y|(V, O, A)) = \left(p([MASK] = \varphi(y))|I_{hybrid}\right) = \frac{\exp(\omega_{\varphi(y)} \cdot h_{mask})}{\sum_{l \in Y} \exp(\omega_{\varphi(l)} \cdot h_{mask})}$$

Where $h_{mask}$ indicates the hidden layer representation of [mask], $w_M$ is the final Layer representations corresponding to $V$, and $M$ can be fine-tuned to minimize the loss in the sample $(V, O, A, y)$ following the rule of cross-entropy. Here are some details different from PVLM

1): Our hybrid pattern contains one or more masks for different templates. For the prediction and metric, we choose only one [MASK]'s result from the probability scores generated for one or more [MASK] tokens through a fusion strategy. The alternative strategy mainly contains: choose which perform best when evaluated to follow and add the probability scores of different [MASK] tokens together directly.

2): Compared with PVLM, when implemented, we filter out the final state for only [MASK] tokens rather than all, which releases the requirement for computational power.

# 4. Experiments

To evaluate the performance of our proposed model on sentence-level MSA and aspect-level MSA, we conducted experiments and evaluations on four datasets (two for sentence-level and two for aspect-level) with a few training sets, comparing our method with the results coming from relevant approaches to multi-modal sentiment analysis.

**Datasets**

|  | Dataset | few | Train | Val | Test |
|---|---|---|---|---|---|
| Aspect-level | Twitter-15 | 36 | 3179 | 1122 | 1037 |
|  | Twitter-17 | 36 | 3562 | 1176 | 1234 |
| Sentence-level | MVSA-M | 96 | 10420 | 1657 | 1657 |
|  | MVSA-S | 36 | 3224 | 403 | 408 |

Table 2. Distribution on the different datasets after splits. "few "represents the balanced random sample of the training set.

1) aspect-based MSA (MABSA): Gathered from Twitter, Twitter-15, and Twitter-17 [29], which are recognized in aspect-level multimodal affective analysis, categorizing emotions as negative, positive, and neutral, and labeling them for supervised training, are widely utilized,

2) sentence-level (MSA): Both MVSA-S and MVSA-D come from [38]. Specifically, MVSA-D is a larger dataset, and it is collected from Twitter and labeled as having three polarities, too.

Firstly, we construct the training set, test set, and validation set from each data set as 8:1:1. To satisfy the requirement of the few-shot task, we gather a few training sets from the original training set by about 1% random sampling. Otherwise, because it's too few for MVSA-S training sets, we build a 2 percent sampling to guarantee the credibility of our approach.

**Setting**

For text modality: consistent with PVLM, we utilize Bert as the pre-raining model to compare. However, Bert is just the backbone of the Bert-series pre-training models, which can employ the prompt-based method and perform better than Bert usually.

For visual modality, we build the presentation vectors from images using NF-ResNet.

For super-parameters, we fine-tune the learning rate from 1e-5 to 1e-4, the embedding length of vision representation from 1 to 5, and the dropout of the Biaffine attention mechanism. See the table for details. Besides, the layers and hidden size of LSTM which are utilized in attention mechanisms and prompt encoder can be changed, too. More importantly, the hybrid pattern has various choices for the set of pseudo-tokens. It changes with the setting of pattern. We set the alternated length for every pseudo-tokens from 1 to 3 and keep the same length of all pseudo-tokens when conduct experiments.

As it has been proved that fine-tuning can lead to instability on different small sets of data, we measure average performance across different training sets gathered by random sampling using the average as a criterion for comparison, and we argue we obtain more stable and robust evaluation results through this strategy.

**Baselines**

For a sufficient comparison, we compare two groups of baselines with our proposed approaches. The first group is the previous tuning, and the main prompt tuning approaches are only based on text modality:

1) BERT [39], as the most popular and competitive pre-train model for sentiment analysis in text-based tasks, where the Masked Language Model is utilized to pre-train the bidirectional transformer to generate deep bidirectional language representation,

2) BERT+BL [29], a variant of BERT, utilizes another transformer encoder layer at the top.

3) Prompt Tuning (PT) [6] only uses single textual prompts, with manual templates as $[s]T < unsed1 < unsed1 > [mask]. [/s]$ or pseudo templates such as $as [s] T\ It\ was\ a\ mask]. [/s]$

4) LM_BFF [4], which prompts tuning with demonstrations like GPT-3 but performs well on the few-shot GLUE benchmark.

The second group is composed of excellent approaches to multi-modal analysis:

1) TOMBERT, baseline on aspect-level MSA, which is based on BERT, and utilizes stack self-attention to capture the interaction information from modalities [29].

2) MMAP [40], the state-of-the-art (SOTA) approach for MABSA, which decreases the gap between image space representation and text space representation.

3) MVAN [25], the SOTA for sentence-level MSA, which adopts an interactive learning mechanism that uses the memory network to model the cross-view dependencies.

4) PVLM, the most competitive and latest method, utilizes prompt learning on text representation and converts the visual representation into an embedding vector with text. It is worth noting that it just trains on a limited training set.

## 5. Main result

**Comparison experiment**

| Model | | Twitter-15 | | Twitter-17 | | MVSA-M | | MVSA-S | |
|---|---|---|---|---|---|---|---|---|---|
| | | Acc | Mac-F1 | Acc | Mac-F1 | Acc | Mac-F1 | Acc | Mac-F1 |
| Text | BERT | 50.81 | 32.8 | 47.91 | 44.18 | 55.73 | 32.56 | 50.29 | 37.91 |
| | BERT+BL | 56.72 | 28.87 | 48.83 | 45.89 | 56.21 | 34.17 | 53.55 | 39.58 |
| | PT (PVLM) | 59.26 | 41.16 | 51.05 | 49.14 | 60.12 | 40.42 | 61.89 | 49.2 |
| | LM-BFF | 59.63 | 35.54 | 49.91 | 48.6 | 60.56 | 40.45 | 58.05 | 39.74 |
| Text+Image | TomBERT | 55.56 | 33.97 | 49.76 | 46.01 | | | | |
| | MMAP | 50.33 | 32.62 | 40.6 | 32.75 | | | - | |
| | MVAN | - | - | - | - | 45.14 | 38.23 | 48.13 | 35.62 |
| | PVLM(hard) | 60.85 | 47.65 | 52.29 | 51.14 | 63.12 | 47.86 | 63.24 | 49.96 |
| | Ours | **62.83** | **54.33** | **53.72** | **54.1** | **64.56** | **49.28** | **65.86** | **57.42** |

Table 3.Comparsion of methods the MVAN and MMAP model just fit the corresponding task ,we fill the result as "-"

In this section, we conduct a comparable experiment on the above datasets among baselines methods and ours. The main result contains:

1) The prompt-based models, PT and LM-BFF, outperform the fine-tuned-based models BERT and BERT-BL in most datasets with little training. This is consistent with previous studies: in NSP-BERT, the use of prompts in few-shot learning significantly improved the results; even with zero sets for training, it performed best on almost NLP tasks. This shows that the use of prompt-based methods for FSA in text mode is reasonable. We speculate that this is mainly because prompting is a natural language way to learn and understand the original input, which makes the prompt-based method less dependent on the training sample.

2) In multimodal, MVAN and MMAP apparently underperform PVLM at the aspect-level and sentence level. Different hard templates make PVLM work best on multiple data sets, but not always. PVLM first introduces prompts into text and image bimodality and proves that prompt-based methods are more effective than SOTA at both sentence and aspect levels separately. This further demonstrates that our prompt-based strategy is suited to few-shot multi-modal sentiment analysis (FMSA). Unfortunately, previous research has shown that in NLP, it is difficult to build a fixed template that performs best in all datasets. As we can see from the results, PVLM does not work best in Twitter 17 when using fixed template 1. This proves that the problem of the fixed template itself still exists under the premise of two-mode, which suggests that a single fixed prompt is not the best solution. This is the basis of our proposed method based on combining the prompts, replacing the single hand-crafted prompt.

3) Our proposed model gets an overall outperformance on the datasets for evaluation. Compared with the best performance of PVLM for accuracy in the Twitter-17, our model still outperforms PLVM by 1.5%. In the rest of the experiments, the best improvement was not more than 2–3%. For the macro-f1, the apparent improvements, from 4% to 8%, perform in the majority of cases. When compared with the single learnable prompt of PLVM, it's mainly due to the fact that our HPM method merges the artificially crafted prompt template with the learnable prompt template, as our method has gained not only the pseudo token but also the fixed prompt template since initialization, which means that both the learnable sequence of prompts and the PrLM model can learn explicit prompt representation from the fixed prompt template to further enhance the performance when the training set is small. Otherwise, different from the p-tuning used by PVLM, we introduce SDPA and biaffine attention to learn interactive information and the underlying dependency syntax representation among learnable prompts. Compared with the hard template method, the primary reason for apparent outperformance is conjectured to be the learnability and adaptability of learnable prompts. Wrapped with our multi-attention-based prompt encoder as an inference unit, the proposed model gathers the syntax and semantic information clearly useful for the reorganization of emotions.

4) We have an interesting discovery that accuracy seems to be unable to be used to measure the effectiveness of the model individually. Sometimes accuracy reaches a relatively high level, even higher than our report, even though macro-f1 is apparently low. Throughout the evaluation process, we pursue the balance of both metrics. Additionally, it can be seen that our model improved more significantly on macrof1 than on accuracy because, in most cases of our experiment, macrof1 still improved significantly with a relatively constant accuracy result.

**Ablation study**

We design ablation experiments among different datasets for both aspect-level tasks and sentence-level tasks to demonstrate every module's necessity. And list the main results in Table $$. Firstly, we remove the biaffine attention (w/o BA), then we further remove the SDPA (w/o SDPA), which means that our reference encoder is just like p-tuning' s, and finally, to evaluate the helpfulness of hybrid hard templates and learnable prompts, we remove the learnable prompt (w/o LP) and another hard template (without DP), respectively. When implementing, we set the pseudo-tokens to be empty and set the length of the pseudo-tokens 0 for w/o LP. The removal of the different modules makes the model's performance ineffective, which indicates that these parts are necessary for our model.

|  | MVSA-S | | twitter-17 | |
| --- | --- | --- | --- | --- |
|  | acc | mac-f1 | acc | mac-f1 |
| w/o BA | 65.21 | 57.15 | 53.01 | 52.27 |
| w/o SDPA | 64.84 | 55.13 | 52.63 | 51.29 |
| w/o DP | 64.53 | 54.24 | 52.01 | 51.43 |
| w/o LP | 63.54 | 53.83 | 51.02 | 50.24 |

Table 4.For BA and SDPA ,the difference between two methods demonstrate the effectiveness of this module ,for DP and LP, which are separately removed ,the necessity of them can be shown between w/o SDPA ,which means remove all attention mechanisms

For BA and SDPA ,the difference between two methods demonstrate the effectiveness of this module ,for DP and LP, which are separately removed ,the necessity of them should be shown between w/o SDPA ,which means remove all attention mechanisms.

From the table we find that: On the one hand, biaffine attention and SDPA have a positive effect on our model, which shows that two attention mechanisms are effective and indispensable for reference encoders to optimize p-tuning in multimodal small sample tasks (FMSA and FMABSA). On the other hand, the simple combined double templates contribute little improvement but a little adverse reduction on macro-f1, which means that a simple mix of two hard templates without learnable prompt optimization may be inapposite.

An overall improvement in the performance using the proposed model can be apparently observed as it successfully uses a combination of continuous template and discrete prompt and we not only enhance the interactive information among modalities by Self-attention and learnable tokens but gather the syntax knowledge by Biaffine attention. In the framework of our model, one or more discrete prompts and learnable prompts are all utilized in a hybrid pattern, whose outputs are fused into natural inputs in each batch for enhancing the syntactic representation of texts. Thus, our proposed framework makes the input of the PLM model richer than the original text. It contains not only discrete prompts used in the PET model but also learnable pseudo-tags used in the p-tuning series.

# 6. Further analysis

**Compare with PVLM in MVSA-single for splits.**

Considering the size of the MVSA-single dataset is too small, to ensure the credibility of the experimental results, a 2% shot was used to compare with PVLM and our method. Since the MVSA-S dataset is seriously imbalanced, we utilize the weighted-f1 together as metrics. As the table demonstrates, we achieve an obvious improvement over all metrics and settings.

| set | 1-shot | | | 2-shot | | |
| --- | --- | --- | --- | --- | --- | --- |
| metric | acc | mac-f1 | weight-f1 | acc | mac-f1 | weight-f1 |
| PLVM | 63.24 | 49.96 | 63.1 | 64.45 | 54.52 | 63.97 |
| Ours | 65.86 | 57.42 | 65.73 | 67.31 | 60.24 | 66.51 |

Table 5. Compare with PLVM

**Does image embedding have positive effects?**

Although we achieve an overall improvement, there are still some details to be considered. First of all, we need to verify that the embedding of image modes is necessary and effective. Although prompt tuning has been used previously to represent text modes, it is necessary to test for redundancy in embedding images for our model.

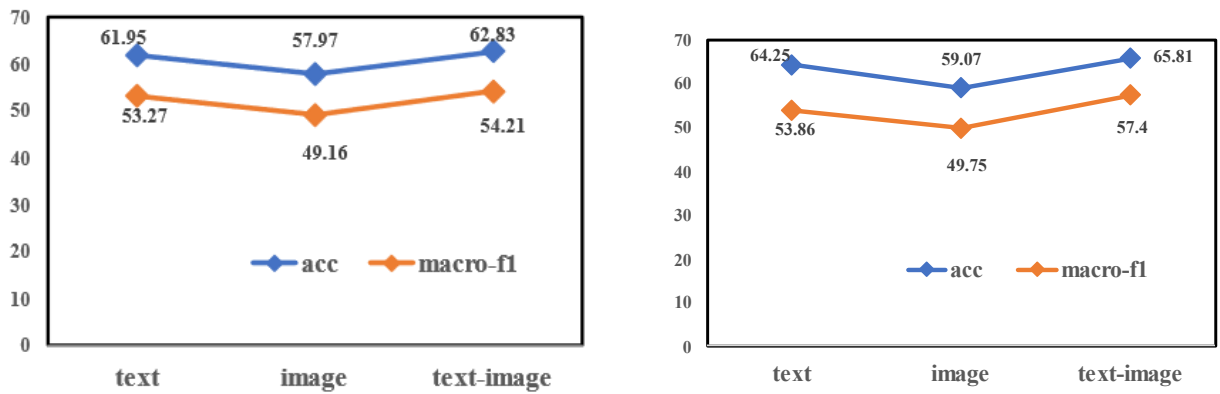

Figure 4. we can find that the experiments based on cross-modalities have the best performance on two levels consistently.

The results from Figure 4 prove that the embedding of images is necessary rather than redundant, and the proposed method is better than tuning on both FSA and FABSA.

**In which modality do learnable prompts work?**

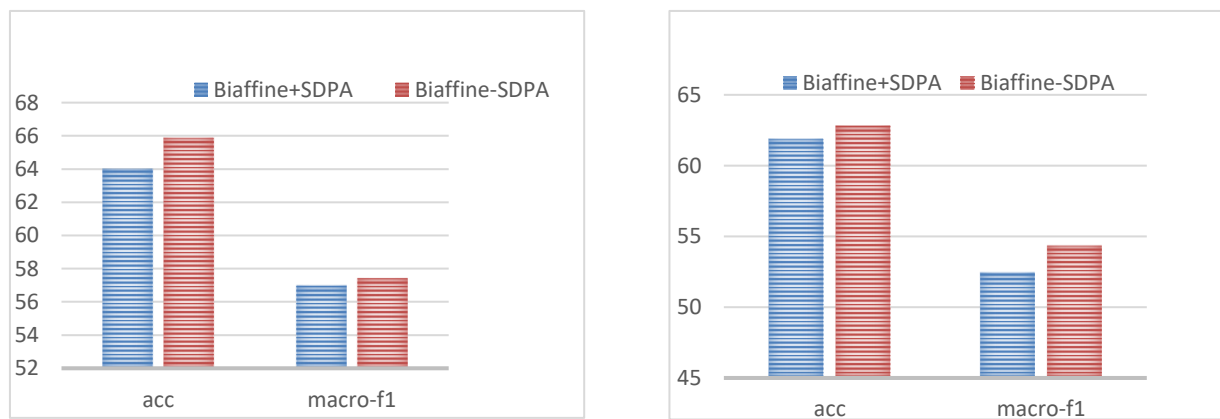

Figure5. we can find that the experiments based on "Biaffine-SDPA" preform better on two levels consistently

To further explore the effect of learnable prompts on the text and visual modalities, we conducted patterns working on the two modalities separately. The test results from Figure $$$ show that the experiment on single mode underperforms that on both, but the performance is still better than that using a simple hard template. It's mainly because our reasoner learns useful cross-modal representation information and narrows the gap between the modalities rather than just depending on the output from a single modality.

**How to fuse the attentions with BI-LSTM efficiently?**

To further explore the appropriate fusion method for the attention mechanisms, we organized experiments on sentence level and aspect-level, respectively. The results from figure $$ show that the best choice is "Biaffine-SDPA", which means using the Biaffine attention to process the linear combination of SDPA output and the original hidden states of BI-LSTM (followed by MLP merging the result of Biaffine attention with the linear combination above mentioned together). "Biaffine +SDPA" denotes the method of directly feeding the hidden state of BI-LSTM into SDPA and Biaffine attention alone. It's mainly because, compared with BI-LSTM's hidden states, the output of SDPA contains extra semantic representation and interaction information, which is helpful to biaffine attention.

In summary, in this chapter, we analyze how our method works between modalities and how to fuse the attention mechanisms better. Compared with the experimental results in single mode, the proposed method is necessary and effective. Through the comparison, we find that the attention mechanism is helpful in integrating semantic information and learning syntax information. A combination of these two attention mechanisms is a better choice than individually using them.

## 7. Conclusion

In this paper, we first propose an inference unit that utilizes fixed templates and learnable prompts together through an inference unit optimized by multi-attention mechanisms. The model applies biaffine attention and SDPA attention in the inference unit to gather syntax and semantic information among learnable tokens, which not only enhances the material for referencing but also achieves sufficient interaction among tokens for two modalities. Moreover, to sufficiently utilize the prompt methods, we conduct patterns for the combination of hard templates and learnable prompt tokens, so our proposed modal gathers not only learnable representation but also natural fixed templates, which are helpful for understanding. Then comparison experiments are conducted on two tasks, followed by the ablation study, the result of which verifies the efficacy of our method and the necessity of the attention mechanisms and learnable prompt method for the hybrid prompt strategy. We further explore and discover how our method benefits from cross-modal interaction instead of a separate modality. We believe our work can inspire creativity in prompt learning and few-shot multimodal sentiment analysis in the future.

# 8. Acknowledgement

The study was funded by the National Natural Science Foundation of China (Grant Nos. 61672144). Zikai Zhou and Haisong Feng contributed equally to this work. Correspondence should be addressed to Baiyou Qiao.